\documentclass{bmvc2k}

\usepackage{amssymb}
\usepackage{soul}
\usepackage{booktabs}
\usepackage{pifont}
\usepackage{comment}

\newcommand{\ie}{\textit{i}.\textit{e}.}

\newcommand{\metricOne}{CoM distance}
\newcommand{\metricOneShort}{CD}
\newcommand{\metricTwo}{Pose Stability Duration}
\newcommand{\metricTwoShort}{PSD}
\graphicspath{{figures/}{BMVC2024_LoKhCo/figures/}}

\title{Measuring Physical Plausibility of 3D Human Poses Using Physics Simulation}

\addauthor{Nathan Louis}{natlouis@umich.edu}{1}
\addauthor{Mahzad Khoshlessan}{mkhoshle@umich.edu}{2}
\addauthor{Jason J. Corso}{jjcorso@umich.edu}{1,2}

\addinstitution{
 Electrical and Computer Engineering\\
 University of Michigan\\
 Ann Arbor, Michigan, USA
}
\addinstitution{
Robotics\\
University of Michigan\\
Ann Arbor, Michigan, USA
}

\runninghead{Louis \etal}{Measure Physical Plausibility Using Simulation}

\def\eg{\emph{e.g}\bmvaOneDot}

\def\etal{\emph{et al}\bmvaOneDot}

\begin{document}

\maketitle

\begin{abstract}
Modeling humans in physical scenes is vital for understanding human-environment interactions for applications involving augmented reality or assessment of human actions from video (\eg sports or physical rehabilitation).
State-of-the-art literature begins with a 3D human pose, from monocular or multiple views, and uses this representation to ground the person within a 3D world space.
While standard metrics for accuracy
capture joint position errors, they do not consider physical plausibility of the 3D pose. This limitation has motivated researchers to propose other metrics evaluating jitter, floor penetration, and unbalanced postures.
Yet, these approaches measure independent instances of errors and are not representative of balance or stability during motion.
In this work, we propose measuring physical plausibility from within physics simulation. 
We introduce two metrics
to capture the physical plausibility and stability of predicted 3D poses from any 3D Human Pose Estimation model. Using physics simulation, we discover correlations with existing plausibility metrics and measuring stability during motion.
We evaluate and compare the performances of two state-of-the-art methods, a multi-view triangulated baseline, and ground truth 3D markers from the Human3.6m dataset.
\end{abstract}

\section{Introduction}
\label{sec:intro}

\begin{figure}[t]
    \begin{center}
    \includegraphics[width=0.85\columnwidth]{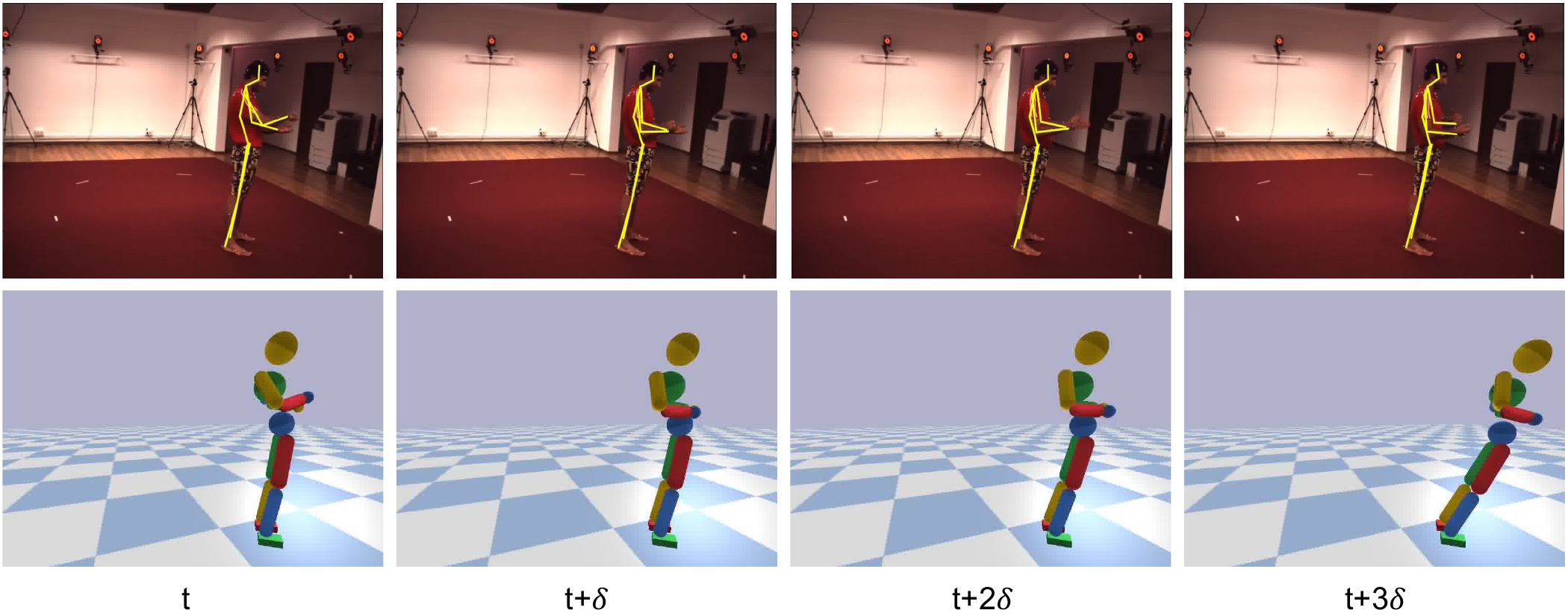}
    \end{center}
    \caption{On the Human3.6M dataset (\textit{S9 - Directions 1}), while the predicted pose (top row) appears plausible, in simulation we see the lean at the hip causes a loss of balance over time.}
    \label{fig:teaser}
\end{figure}

Physically grounding objects from video is vital for understanding spatial relationships \cite{weng2021holistic}, geometric properties~\cite{tulsiani2018factoring}, and contact forces \cite{brubaker2009estimating, louis2022learning}.
For humans, this allows us to understand how a person interacts within an environment and models their actions for entertainment or evaluation in sports and physical rehabilitation~\cite{fieraru2021aifit,patrona2018motion,monzani2000using}.
To physically ground a person from video, common learning-based approaches employ state-of-the-art 3D human pose estimation~(HPE) methods \cite{kocabas2020vibe,zheng20213d,zhang2021direct,pavllo:videopose3d:2019} and measure performance using Mean Per Joint Position Error (MPJPE), 
Mean Per-vertex Error (MPVE), 
or Probability of Correct Keypoint (PCK).   
But even impressive pose estimation progress on computer vision datasets \cite{ionescu2013human3, vonMarcard2018, deepcap} present notable flaws in pose estimates, such as floating, foot-skating, ground penetration and unnatural positions \cite{rempe2020contact}.
Moreover, MPJPE---the primary quantitative metric---often displays little correlation between low error and visual quality of results~\cite{kanazawa2018end}.
These discrepancies led us to question the physical plausibility, hence physical grounding, of these predicted 3D human poses. 
Prior studies \cite{reitsma2003perceptual, hoyet2012push} have shown that humans are sensitive to motions lacking perceptual realism which may negatively impact applications involving augmented or virtual reality and physical understanding.

Identifying these issues, researchers have explored physics-based metrics \cite{rempe2020contact,yuan2021simpoe,shimada2020physcap,gartner2022trajectory, kocabas2020vibe,tripathi20233d} for 3D HPE. For measuring realistic contact with the ground, metrics such as footskate \cite{rempe2020contact}, foot slide \cite{yuan2021simpoe}, and ground penetration \cite{yuan2021simpoe, shimada2020physcap} have been proposed.
To highlight unnatural jittering, others have computed smoothness losses from the velocity or acceleration of joints \cite{shimada2020physcap, yuan2021simpoe, gartner2022trajectory, kocabas2020vibe}. And for stationary stability, Shimada \etal \cite{shimada2020physcap} and Tripathi \etal \cite{tripathi20233d} introduced terms to measure balanced and unbalanced static postures.
However, these approaches do not measure stability during motion or over time, rather independent instances of errors only.
In Figure \ref{fig:teaser}, we show how an unnatural lean in the predicted pose may appear stable but overtime will lead to a loss of balance in simulation.

To overcome these limitations, we evaluate the physical plausibility of 3D HPEs through physics simulation. 
Physically plausible poses should obey the laws of physics, 
while physical simulators must replicate friction, gravity, and collisions.
We hypothesize a link between pose plausibility and simulation stability: more stable simulations likely indicate more plausible poses.
We measure the physical plausibility on simulated bodies using two new metrics: \metricOne, the center-of-mass trajectory distance with the kinematics reference,
and \metricTwo, the time at which a pose can be simulated before reaching an irrecoverable failure. These two measures describe the temporal stability of the simulated body undergoing motion, and unlike earlier physics-based metrics, we encapsulate dynamics and collisions without relying on hand-crafted heuristics.
We also note, our simulated-based metrics can be applied to any off-the-shelf 3D HPE result without ground truth data.

Our contributions are two physical simulation-based metrics, \metricOne\ and \metricTwo, for evaluating the physical plausibility of 3D HPEs and a comprehensive analysis of publicly available state-of-the-art methods and baselines.
All simulation and evaluation code have been made publicly available \footnote{\url{https://github.com/MichiganCOG/Simulation_Physical_Plausibility}}.

\section{Related Work}

\subsection*{3D Human Pose Estimation}
3D HPE is performed using monocular \cite{kocabas2020vibe, ROMP, pavllo:videopose3d:2019} or multi-view \cite{remelli2020lightweight,pavlakos2017harvesting,iskakov2019learnable,qiu2019cross, zhang2021direct, tu2020voxelpose} camera inputs.
Monocular camera-based solutions are cost-effective, however, they present unique challenges due to their inherent depth ambiguity.
Multi-view methods resolve these ambiguities through aggregation of multiple camera but require known camera projection matrices.
Within these two classes, single-stage approaches approximate the 3D pose by estimating parameters of a statistical body model \cite{loper2015smpl, xu2020ghum, kocabas2020vibe} and two-stage methods \cite{pavllo:videopose3d:2019,zheng20213d,pavlakos2017harvesting} employ a 2D detector to predict a 2D pose and ``lift'' it into 3D.
The most commonly used metrics for quantitative evaluation, MPJPE, MPVE, and PCK, essentially measure the average displacement between predicted joints and corresponding ground truth joints. 
However, they fall short in their ability to quantitatively evaluate plausible postures or motion. Our proposed metrics \metricOne\ and \metricTwo\ address this through physical simulation of the predicted pose.

\subsection*{Physics-aware 3D Human Pose Estimation}
The prior 3D HPE methods are kinematics-based, only modeling joint location and disregard underlying forces and environmental factors. 
Failure to account for dynamics results in errors such as floating, foot skating, and unrealistic postures \cite{rempe2020contact}.
To address these implausibilities, some works have proposed ground contact metrics to measure ground penetration \cite{yuan2021simpoe, shimada2020physcap} and foot sliding (or skating) \cite{rempe2020contact, yuan2021simpoe}.
But these metrics use handcrafted heuristics and are most reliable on a calibrated floor plane.
To measure stability of poses, other works \cite{shimada2020physcap,tripathi20233d} incorporate terms from biomechanics literature \cite{hof2005condition,hof2007equations,winter1995human} to discern balanced stationary postures.
However, these functions are only valid for stationary poses and not a pose undergoing motion because it does not account for the velocity of the center of gravity \cite{hof2005condition}.
Despite these advances, these metrics do not analyze the temporal impacts of physical implausibility or a body in motion. For a more comprehensive evaluation, we look to physical simulation as a test bed for plausible motion assessment under the effects of gravity, frictional forces, and collisions.

\section{Method}
\begin{figure*}[t]
    \centering
    \includegraphics[width=0.99\textwidth]{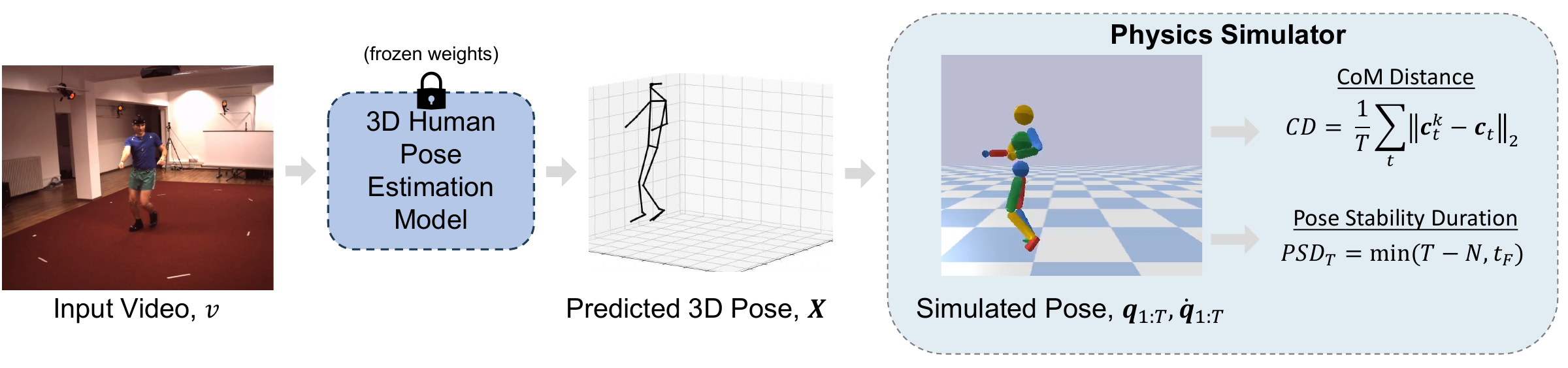}
    \caption{From a video $v$, we estimate 3D human poses $\mathbf{X}$ from an off-the-shelf 3D HPE model. Next, we initialize kinematic joint targets $\mathbf{q}^k_{1:T}$ on a simulated body and optimize it to mimic the reference motion under simulated environmental effects. We measure the plausibility of this optimized output using \metricOne\ and \metricTwo.}
    \label{fig:main}
\end{figure*}

We propose a new evaluation method for assessing the physical plausibility of 3D human pose estimates (HPE);  
we show an overview of our approach in Figure \ref{fig:main}. 
From a given video, we predict 3D human poses using an off-the-shelf 3D HPE model and estimate a reference kinematic trajectory. 
Following, we apply these kinematics to a simulated body and optimize its motion under physical effects of the simulation environment. 
Finally, we measure the stability of the simulation using our proposed metrics to quantitatively evaluate the physical plausibility of the predicted 3D poses. In Supplemental Material, we discuss qualitative analysis of estimated ground contacts forces.

\subsection{Kinematic Initialization} \label{sec:kinematic_init}
Given a video $v$ of a person performing some action, we predict a sequence of 3D skeletal poses $\mathbf{X} \in \mathbb{R}^{T \times J \times 3}$ using an off-the-shelf 3D HPE method, for $J$ joints across $T$ frames. Models generating a mesh, such as SMPL \cite{loper2015smpl}, can use a pre-trained regressor to estimate a 3D skeletal pose.
We apply minimal pre-processing to reduce noise in $X$, 
first a median filter ($w=15$ frames) and then constraining bone lengths to their mean values.

To apply the pose to a simulated body, we initialize a kinematic representation using a set of kinematic poses $\mathbf{q}^k_{1:T}$ from $X$. 
Each kinematic pose $\mathbf{q} = (q_1, q_2, ..., q_D)$ is a concatenation of all joint angles, parameterized as quaternions, for $D$ degrees-of-freedom (DOF) on the simulated body.
The kinematic pose is generally optimized from an inverse kinematics problem,
however the ill-posed nature of this optimization often produces implausible and substantially different poses from the predictions in $X$.
Similar to~\cite{peng2018deepmimic}, we define a kinematic tree and use change of basis rotations, from the root to the end effectors, to approximate the joint angles in $\mathbf{q}$. 
This assumes both the skeletal pose and the simulated body have compatible kinematic trees.
The resulting output, $\mathbf{q}^k_{1:T}$, is a direct kinematic representation of the 3D pose sequence without alteration from an inverse kinematics solver.

The simulated body is a humanoid
with $28$ DOFs, 
and $12$ controllable joints. This includes eight spherical joints with $3$ DOFs each and four revolute joints (knees and elbows) with $1$ DOF each.
The root node (pelvis) is excluded as a controllable joint, since applying forces directly to the root of the kinematic tree is not realistic and lacks physical meaning \cite{brubaker2009estimating}.
Each joint is paired with a Stable PD controller \cite{tan2011stable}, which takes as input target joint angles, joint velocities, and gains ($k_p$,$k_d$) to output a torque on each joint.
At each time step, the torque inputs are computed as 
\begin{equation}
    \mathbf{\tau} = k_p (\mathbf{q}^k - \mathbf{q}) + k_d (\dot{\mathbf{q}}^k - \dot{\mathbf{q}}).
\end{equation}
Here, $\mathbf{q}^k$ represents the target kinematic pose while $\mathbf{q}$ is the current kinematic pose, and joint velocities $\dot{\mathbf{q}}^k_{1:T}$ are estimated using first-order finite differences. The proportional and derivative gains $k_p$ and $k_d$ are fixed across all experiments.

\subsection{Trajectory Optimization}
We perform trajectory optimization \cite{al2012trajectory,al2018robust,gartner2022trajectory} to optimize the kinematic pose targets so that the simulated body mimics the action in video.
Given that physical simulators are highly advanced and generally non-differentiable, we use the derivative-free Covariance Matrix Adaptation Evolution Strategy (CMA-ES) \cite{hansen2006cma} algorithm. We run the simulator at $1$kHz with kinematic targets set at $25$Hz, optimizing for $200$ iterations and a population size of $100$, on eight overlapping windows of length $0.5$s.
Rather than optimize all time steps simultaneously, we reduce the search space by representing the targets as Euler angles in a cubic B-spline and optimize the knots.
We minimize the following cost functions:
The center-of-mass loss 
,
\begin{align}
    L_{COM} = \sum_t \left( \mathbf{c}^k_t - \mathbf{c}_t \right)^2,
\label{eq:loss_com}
\end{align} 
the distance between the COM of the kinematic target, $\mathbf{c}^k$, and the current COM, $\mathbf{c}$, from the simulated body.
The center-of-mass velocity loss,
\begin{align}
    L_{COMv} = \sum_t \left( \dot{\mathbf{c}}^k_t - \dot{\mathbf{c}}_t \right)^2,
\label{eq:loss_comv}
\end{align} 
the distance between COM velocities of the kinematic target and the current velocity.
The root orientation loss,
\begin{align}
    L_{orn} = \sum_t \arccos\left( \left| \left< \mathbf{q}^k_{root}, \mathbf{q}_{root} \right> \right| \right),
\label{eq:loss_orn}
\end{align} 
constraining the root orientation of the body to align with the kinematic reference.
The kinematic pose loss,
\begin{align}
    L_{pose} = \sum_{t} W * \left(\mathbf{q}^k_{t} - \mathbf{q}_{t} \right)^2,
\label{eq:loss_kin}
\end{align} 
and the kinematic pose velocity loss,
\begin{equation}
    L_{vel} = \sum_{t} W * \left(\dot{\mathbf{q}}^k_{t} - \dot{\mathbf{q}}_{t} \right)^2,
\label{eq:loss_kinv}
\end{equation} 
minimizing the joint angle and joint velocity differences. We minimize these two losses by parameterizing the joint angles as Euler angles and $W$ is the joint weighting array applied element-wise to each joint.
Next, is the joint acceleration loss,
\begin{equation}
    L_{acc} = \sum_t \|\ddot{\mathbf{q}_t}\|^2.
\label{eq:loss_acc}
\end{equation} 
to encourage a smoother trajectory and avoid jittery motions.
And finally, we introduce a contact loss,
\begin{equation}
    L_{feet} = \sum_t \| \mathbf{p}_t^k - \mathbf{p}_t \|_1,
\label{eq:loss_foot}
\end{equation} 
to encourage alignment of foot contact states with those of the input pose.
We use joint values in $X$ to estimate ground truth foot contacts $\mathbf{p}^k$, discussed briefly in Supplemental Material.
The foot contact states in simulation $\mathbf{p}$ are determined using a small height threshold ($0.0005$) between the ground plane and feet.
We observe that this constraint helps to maintain balance when shifting weight between feet. 
The total optimization objective function is a linear combination of the aforementioned losses with the following weights $w_{COM}=20$, $w_{COMv}=0.5$, $w_{orn}=1.0$, $w_{pose}=1.0$, $w_{vel}=5e^{-3}$, $w_{acc}=1e^{-10}$, and $w_{feet}=1.5$.

\subsection{Simulation-based Metrics}
\label{sec:metrics}
Finally, we measure the stability of the optimized simulated body motion using our proposed metrics \metricOne\ and \metricTwo\ to correlate to physical plausibility.

\paragraph{Metric 1: \metricOne}
The \metricOne\ is straightforward, computing the L2 distance between the kinematic COM trajectory and the final optimized COM,
\begin{equation}
    \text{\metricOneShort} = \frac{1}{T}\sum_t\|\mathbf{c}^k_t - \mathbf{c}_t\|_2,
\label{eq:metric_com_distance}
\end{equation} 
in millimeters (mm).
Comparing the final trajectory with the kinematic reference indicates how well the 3D pose estimate $X$ can be simulated. A low value suggests that the reference pose is plausible and easy for the simulated body to follow, while a high value indicates significant deviations from the reference.

\paragraph{Metric 2: \metricTwo}
We introduce \metricTwo\ to understand the time at which instability in simulation occurs. 
First, we identify two states of motion from the predicted pose: stationary and non-stationary.
Values $|\dot{\mathbf{c}}^k_t|$ below a threshold ($250$mm/s) are considered stationary, while values above that threshold are considered non-stationary.

For the stationary state, biomechanics literature \cite{hof2005condition,hof2007equations,winter1995human} states that a stationary pose is considered balanced if its center of gravity (CoG) falls within its base-of-support (BoS), \ie\ the convex hull of all ground contacts.
We compute the BoS on the simulated body, and denote the number of instances where the center of gravity is beyond the convex hull with $N$.
A slight loss of balance in the stationary case is recoverable.
For the non-stationary state, a body in motion may have its CoG lie outside of its BoS, invalidating the previous assumption~\cite{hof2005condition}.
Within the simulator we note a loss of balance in this state as the time $t_F$ when a body part, other than the foot, from the simulated body touches the ground. A loss of balance in the non-stationary case is not recoverable.
We note that prior knowledge of the action (\eg\ cartwheels, push-ups) may be required for accurate computation.
We defined this metric as
\begin{equation}
    \text{\metricTwoShort}_T = \min(T - N, t_F),
\end{equation}
where $T$ is the total number of frames in each sequence.
By examining both stationary and non-stationary poses, we assess the maximum number of simulated frames before a catastrophic failure, \ie\ an unrecoverable deviation from the reference motion.

\section{Experiments}
\subsection{Evaluation Dataset}
All experiments are performed on Human3.6M \cite{ionescu2013human3}, a video dataset capturing human actions using four cameras and a motion capture system where full camera projection matrices are assumed known.
We use the validation subjects \textit{S9, S11}, and the same subset of actions as prior work \cite{shimada2020physcap}: \textit{Directions, Discussion, Greeting, Posing, Purchases, Photo, Waiting, WalkDog, WalkTogether}, and \textit{Walking}.
Videos are downsampled from $50$fps to $25$fps on $100$ frames for computational efficiency in simulation.

\subsection{Evaluation Models}
We evaluate physical plausibility using a monocular model, PoseFormer \cite{zheng20213d}, a physics-aware model, NeuralPhysCap \cite{shimada2021neural}, and a multi-view baseline.
PoseFormer is a spatio-temporal transformer that produces a 3D human pose from 2D pose detections. 
PoseFormer outputs a pose in the camera space, hence to estimate a global pose we minimize the 2D reprojection loss through a differentiable projection function. 
NeuralPhysCap proposes a differentiable framework to generate physically plausible poses through contact estimation, force estimation, and physics-aware optimization.
We chose both PoseFormer and NeuralPhysCap for their publicly available code and to cover two kinematic and physics-aware aspects of monocular 3D HPE models.
Last, we include a multi-view triangulated baseline that generates global 3D poses by applying RANSAC triangulation on 2D wholebody MSCOCO detections \cite{lin2014microsoft,jin2020whole} from all available camera views. 
NeuralPhysCap generates a $16$-joint skeleton and the multi-view baseline produces a $23$-joint skeleton, including heel and toe joints, differing from the $17$-joint H36M skeleton. However, we only compute MPJPE on the mutual joints. We provide additional details on pose formats in Supplemental Material.

\subsection{Evaluation Metrics} \label{sec:evaluation_metrics}
We evaluate models using kinematics-based and physics-based metrics. 
First, we use MPJPE, measured in mm, to compute the pose error relative to the root location.
Then, MPJPE-2D to evaluate the 2D image alignment between the projected ground truth and projected 3D prediction. And MPJPE-G to account for the global 3D position error in the world space.
For physics-based metrics, in addition to our proposed metrics, we include Footskate (FS\%) \cite{rempe2020contact} and ground penetration (GP) \cite{yuan2021simpoe, shimada2020physcap}.
FS\% measures the percentage of frames where the foot moves more than $2$cm while in contact with the ground.
And GP computes the average distance to the ground for joints below the ground plane.
We discuss details on estimating floor height and ground contact states in Supplemental Material.

\section{Results}
\begin{figure*}
    \centering
    \includegraphics[width=0.85\textwidth]{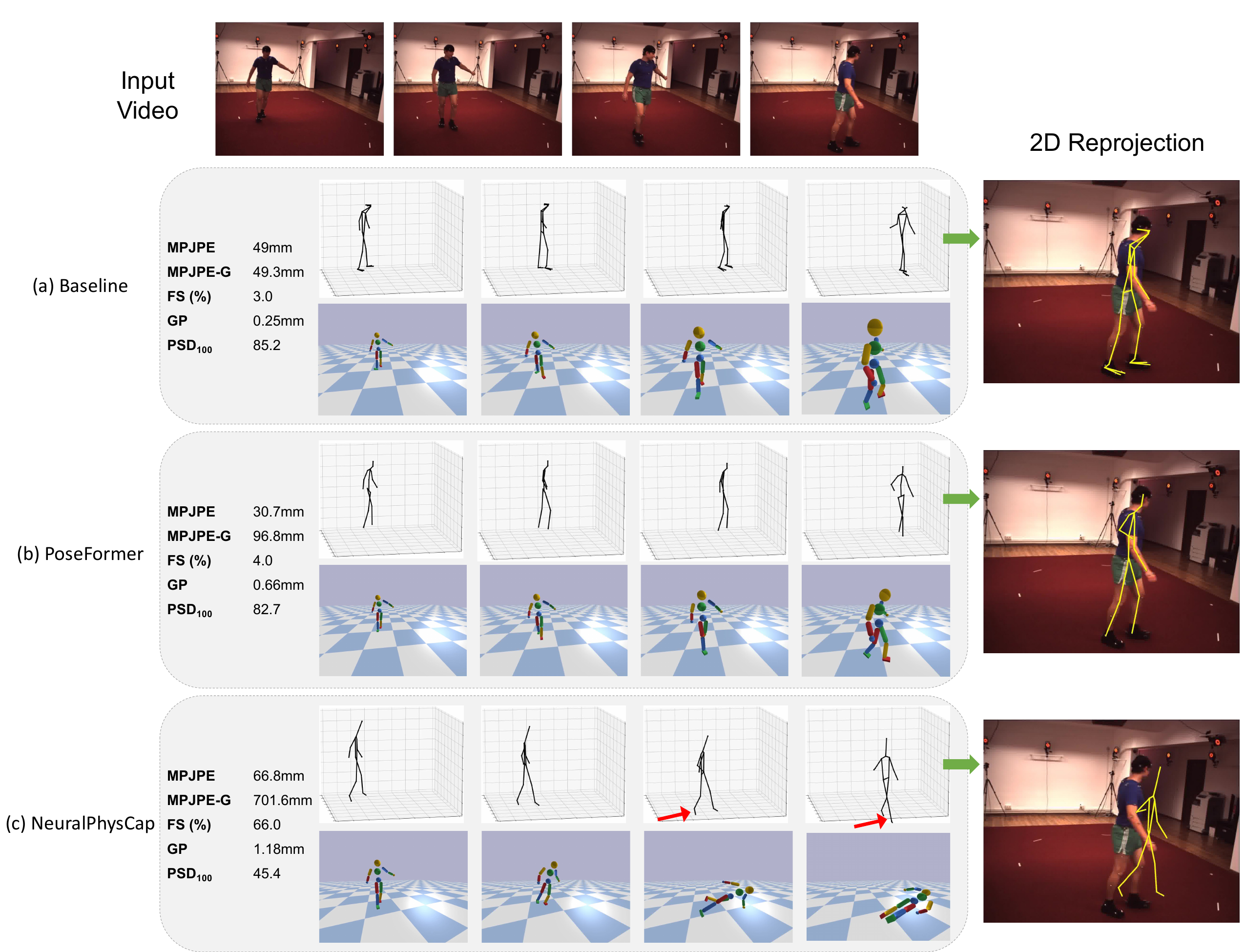}
    \caption{For the \textit{S11 - WalkTogether} example, we show the 3D pose, optimized simulation, and 2D reprojection. Inaccurate camera and ground plane assumptions in (c) causes the motion to fail early on as the simulated body tries to step through the ground plane (red arrows).}
    \label{fig:qual}
\end{figure*}

\begin{figure}
    \centering
    \includegraphics[width=0.75\textwidth]{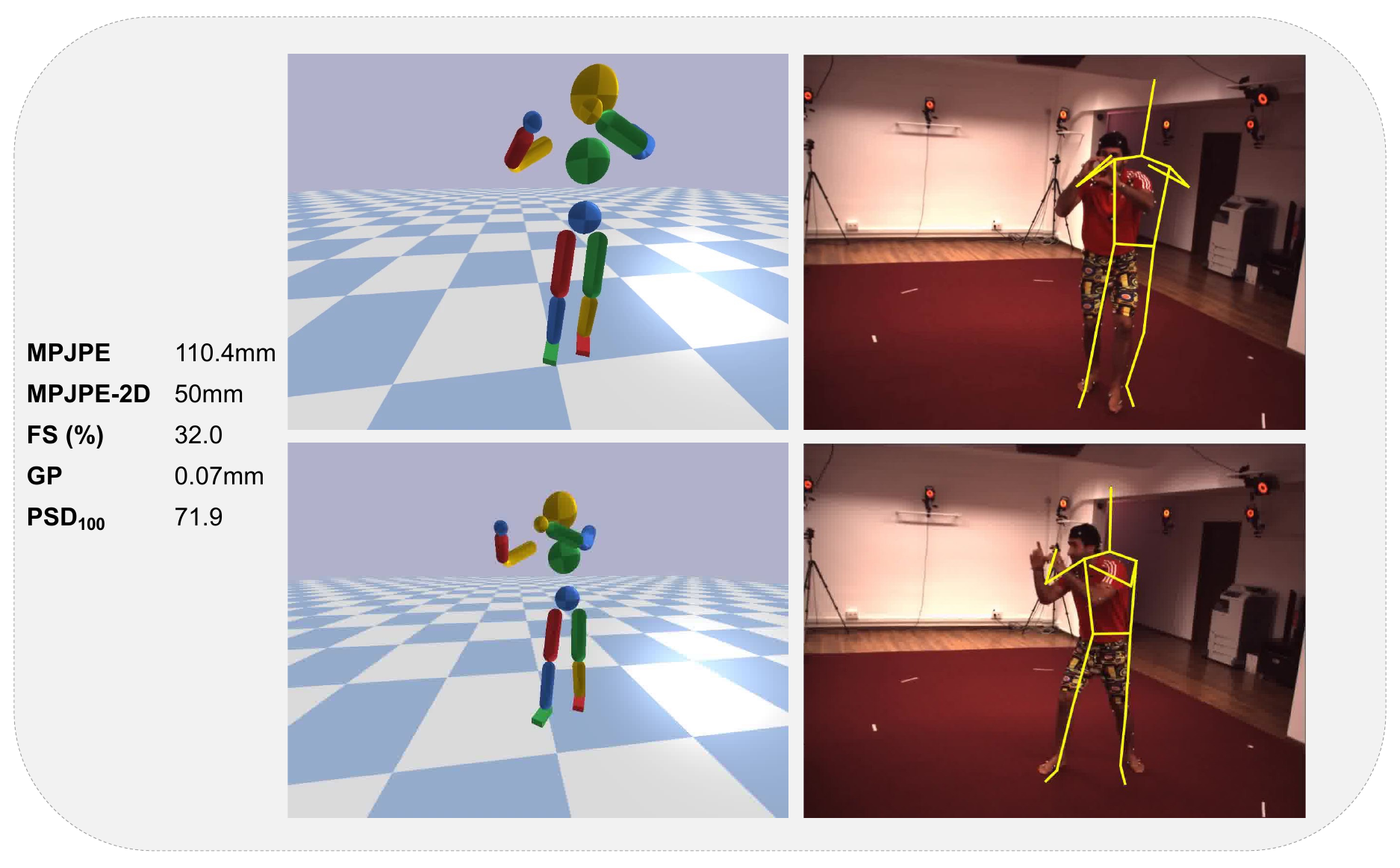}
    \caption{For \textit{S9 - Photo 1} example, we show a misaligned 2D re-projected (right column) can still produce physically plausible simulation (left column) from NeuralPhysCap \cite{shimada2021neural}.  While this example displays higher MPJPE-2D=$50.0$mm, we measure \metricTwo\ to be on par with other methods, $\text{\metricTwoShort}_{100}=71.9$.}
    \label{fig:2d_image_alignment}
\end{figure}

In our main results in Table \ref{tab:results_val_h36m}, we introduce GT 3D, the ground truth surface markers from H36M, as a soft upperbound for the 3D pose. 
The first three columns focus on MPJPE-derived metrics which emphasize spatial alignment of poses, while the latter columns evaluate physical plausibility.
PoseFormer~\cite{zheng20213d} has the lowest MPJPE, $42.5$mm, and MPJPE-2D, $10.6$ pixels.
The baseline has the lowest MPJPE-G with $57.2$mm because it utilizes multiple views and known camera projection matrices to regress a global pose. 
NeuralPhysCap has the highest amount of error, likely due to its miscalculation of the camera intrinsic parameters. This can be seen in the 2D spatial alignment in Figures \ref{fig:qual} and \ref{fig:2d_image_alignment}, however we observe less impact in our proposed plausibility metrics.

\begin{table*}[ht]
    \centering
    \resizebox{0.95\textwidth}{!}{%
    \begin{tabular}{l | l l l l l l l}
    \toprule
    Method & MPJPE $\downarrow$ & MPJPE-G $\downarrow$ & MPJPE-2D $\downarrow$ & FS (\%) $\downarrow$ & GP $\downarrow$ & \metricOneShort\ (Ours) $\downarrow$ & \metricTwoShort$_{100}$ (Ours) $\uparrow$ \\
    \midrule 
         GT 3D & - & - & - & \textbf{0.0} & 1.08 & 33.7 & 63.1 \\ 
         NeuralPhysCap \cite{shimada2021neural} & 81.6 & 439.6 & 36.3 & 29.7 & 2.62 & 29.6 & 62.3 \\
         PoseFormer \cite{zheng20213d} & \textbf{42.5} & 299.4 & \textbf{10.6} & 4.4 & 0.30 & 36.2 & 64.8 \\
         Baseline & 55.6 & \textbf{57.2} & 12.0 & 2.6 & \textbf{0.26} & \textbf{27.3} & \textbf{70.6} \\ 
    \bottomrule
    \end{tabular}
    }
    \caption{Results on the validation subset of the Human3.6M dataset, comparing kinematics and physics-based plausibility metrics.}
    \label{tab:results_val_h36m}
\end{table*}


In the next four columns we examine physical plausibility. The baseline produces the second lowest foot skate, $2.6\%$ and the lowest GP error, $0.26$mm, demonstrating a consistent estimation of pose with ground plane.
While GT 3D has no foot skate error, it surprisingly has the second highest ground penetration. But this variance in ground penetration depth could be due to the inherent margin of error from motion capture surface markers.
Comparatively with our metrics, the baseline is demonstrated to be the most physically plausible with the lowest \metricOneShort, $27.3$mm, and the highest \metricTwoShort$_{100}$, $70.6$ frames.
Unlike the prior metrics we observe that NeuralPhysCap performs closer to the other methods within simulation, with $29.6$mm and $62.3$ frames, respectively.
This indicates some invariance to strict 2D spatial alignment of the pose prediction and an increased focus on the plausibility of its motion.
In Table \ref{tab:results_per_class}, we examine the per-class performance of \metricTwo.
The lowest performing classes: \textit{Greeting, Purchases, Waiting, and WalkDog}, contain crouching and bending over movements which increases the difficulty of optimizing balanced movements on the simulated body.

\begin{table*}[ht]
    \centering
    \resizebox{0.95\textwidth}{!}{%
    \begin{tabular}{l | l l l l l l l l l l | l}
    \toprule
    Method & Dir. & Disc. & Greet & Photo & Pose & Purch. & Wait & WalkD. & WalkT. & Walk & Avg. \\ 
    \midrule
    GT 3D &  \textbf{91.6}&  \textbf{83.8}&  10.5&  \textbf{94.5}&  \textbf{82.7}&  37.3&  37.6&  36.9& \textbf{80.0}&76.1&63.1\\
    NeuralPhysCap \cite{shimada2021neural} &  86.5&  82.3&  73.1&  89.6&  62.7&  35.8&  \textbf{70.4}&  52.1& 26.2&44.3&62.3\\
    PoseFormer \cite{zheng20213d} &  69.8&  77.8&  \textbf{81.0}&  82.8&  80.5&  13.3&  42.2&  \textbf{55.8}& 66.1&\textbf{78.3}&64.8\\
    Baseline &  88.2&  82.4&  66.0&  87.6&  72.9&  \textbf{72.4}&  53.5&  27.7& 77.8&77.2&\textbf{70.6}\\
    \bottomrule
    \end{tabular}
    }
    \caption{We break down the per-class performance for \metricTwo\ ($\text{\metricTwoShort}_{100}$).}
    \label{tab:results_per_class}
\end{table*}


 We provide a visual example in Figure \ref{fig:qual}. The baseline (a) and PoseFormer (b) show comparable performance on physical plausibility metrics and produce similar simulation results. While, NeuralPhysCap fails early due to inaccurate camera intrinsic and ground plane estimation, more notable in this example due to the large amount of motion for this action. 
The red arrows show where the 3D pose attempts to penetrate the ground plane and on the right column, we see the 2D misalignment in the image plane 

\textbf{2D Image Alignment} \quad
In Figure \ref{fig:2d_image_alignment}, we provide an example of NeuralPhysCap to show how spatially misaligned poses can still generate physically plausible poses.
Although the kinematic metrics errors are higher than average, $110.4$mm for MPJPE and $50.0$mm for MPJPE-2D, physical plausibility according to our metrics \metricOneShort, $0.381$, and \metricTwoShort$_{100}$, $71.9$, are on par with Table $\ref{tab:results_val_h36m}$.
Conversely, FS, $32\%$, is very high while and GP is $0.07$mm. NeuralPhysCap explicitly corrects poses for physical implausibilities, so for relatively little motion it makes sense that the output pose is plausible, despite poor spatial alignment.
This suggests that we are assessing the plausibility of the pose and not strict pose alignment, which can improved with more accurate camera parameter estimation.

\section{Discussion}

In this work, we propose two simulation-based metrics, \metricOne\ and \metricTwo, to measure the physical plausibility of 3D HPE using physical simulation. \metricOne\ captures how closely the 3D pose estimate can be simulated when used as a kinematic reference. And \metricTwo\ records the time the pose can be simulated before reaching an irrecoverable failure.
While prior approaches capture independent instances of physical implausibilities, they do not demonstrate the ability to understand the progression of these instabilities.
Our metrics incorporates the temporal aspects of stability while also integrating simulated physical properties. Moreover, we demonstrate some invariance to strict spatial alignment of 3D poses, highlighting instead the feasibility as shown by 2D image alignment disparities.
Our experiments shows consistency with other kinematics and physics-based metrics when determining the most physically plausible output.

\textbf{Limitations} \quad 
This work is impacted by the convergence of the trajectory optimization. If the optimization falls into a local minimum, it may deviate from the reference motion which we mitigate through our constraints.
Additionally, we retarget all humans to the same simulated body, which can introduce modeling errors when the shape and size vary drastically between subjects. Future work may resolve this by modifying the limbs of the simulated body to reflect the approximate shape and size attributes of the detected humans.

\section{Acknowledgements}
This work has been supported by NIH Award 5 R01 HL-146619-05.

\bibliography{egbib}
\end{document}


\maketitle

\section{Supplemental Material}
\subsection{Contact Force Estimation}
\begin{figure}[ht]
    \centering
    \includegraphics[width=0.85\textwidth]{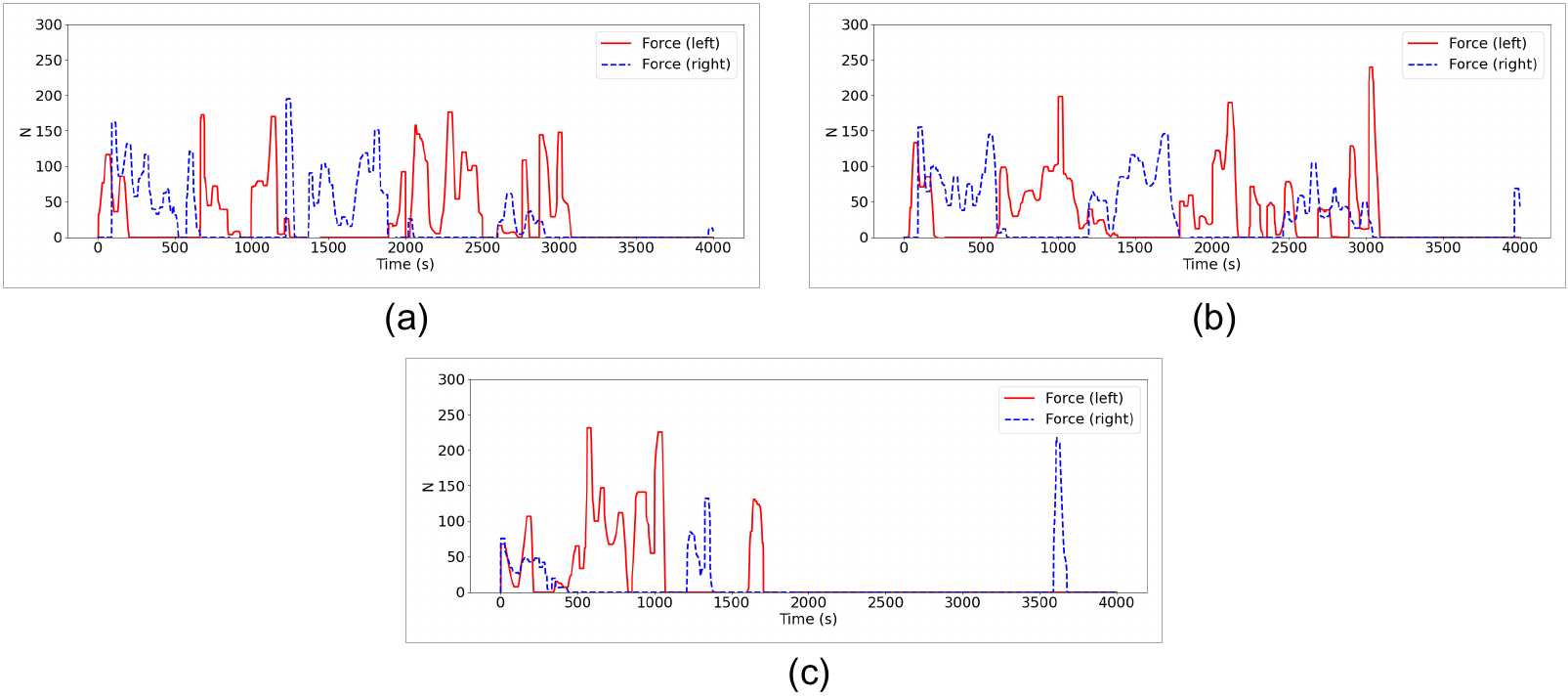}
    \caption{Generated ground contact forces on \textit{S11 - WalkTogether 1} from Human3.6m. The following results are from (a) Baseline (b) PoseFormer (c) NeuralPhysCap}
    \label{fig:contact_forces}
\end{figure}

We demonstrate the ability to generate plausible ground contact forces from within the simulator in Figure \ref{fig:contact_forces} for the \textit{S11 - WalkTogether 1} sequence.
Because there are no ground truth contacts or forces, this is useful for subjective analysis. The ground contact states are estimated from the ground truth MoCap markers and the contact forces are approximated within the physical simulator by summing the normal and lateral forces when the foot joint makes contact with the ground plane.
For the walking motion, we expect alternating peak magnitudes for each foot and we see that the Baseline and PoseFormer show this in the estimated forces and ground contacts. While the NeuralPhysCap example fails very early on.

We determine the height of the ground plane from the average $\lfloor 0.05 \ast T\rfloor$ lowest joint locations, roughly $5\%$ of the $T$ frames. The ground plane is assumed to be normal to the initial pose location.
And to estimate ground contact states on each sequence, we use the estimated ground plane and follow the same heuristics as Rempe \etal\ \cite{rempe2020contact} employing a height threshold of $5$cm and velocity threshold of $2$cm/s on the foot joints to identify ground contact.

\subsection{Skeleton Pose Formats}
\begin{figure}[ht]
    \centering
    \includegraphics[width=0.8\textwidth]{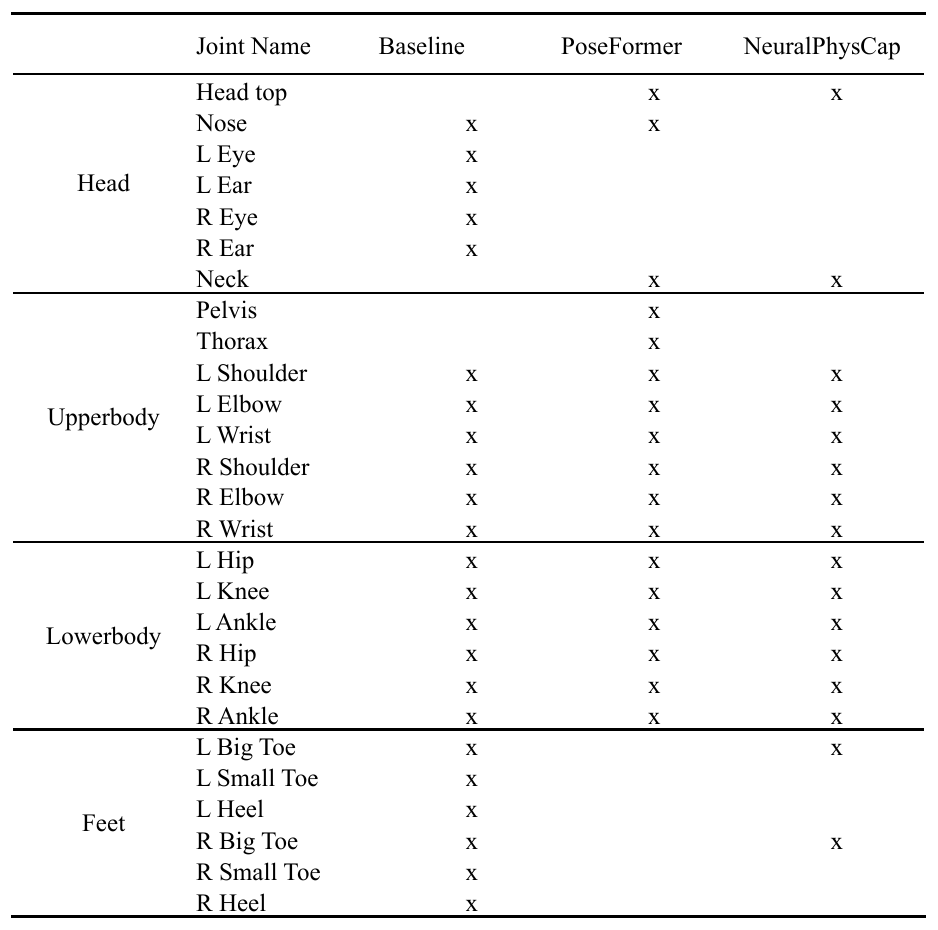}
     \caption{The skeleton pose formats and the shared joints between the different HPE methods. If the pelvis joint is not detected, it is estimated using the midpoint of the left and right hip joints. We apply a similar approach to approximate the positions of the neck and thorax.}
     \label{tab:supp_all_joints}
\end{figure}

In Figure \ref{tab:supp_all_joints}, we layout the joints supported by each skeletal pose.
PoseFormer uses the H36M skeleton which is composed of markers on top of the skin rather than body joint centers.
The Baseline and NeuralPhysCap use skeleton formats from human annotated datasets, \eg MSCOCO, with annotated joint centers.
All of the skeleton poses are mapped to the same simulated body using in-common upperbody and lowerbody joints. 
The joint angles for the simulated body are neck, chest, shoulders (2), pelvis, elbows (2), hips (2), knees (2), ankles (2).
There are a total of $12$ controllable joints on the simulated body, the pelvis is not controlled.

We define a kinematic tree and use change of basis rotations, from the root (pelvis) to the end effectors (hands and feet), to approximate all joint angles.
First, we define the origin basis for the coordinate system $A = [\mathbf{x},\mathbf{y},\mathbf{z}]$.
Then, we find the orientation of the pelvis $B = [\mathbf{x},\mathbf{y},\mathbf{z}]$ from the positions of the thorax, pelvis, and right hip using the right hand rule. 
And finally, the root orientation for the pelvis is defined as the rotation between $A$ and $B$. We repeat this for the next pair of joints (\eg\ pelvis and chest), traversing outwards to the end effectors for the remaining joints.
If toe and heel joints are not detected, the orientation of the ankle for the kinematic initialization is unknown. Instead, we initialize the ankle with a neutral pose and impose no constraints during the optimization process.

\subsubsection*{Impact of toe and heel joints}
\label{sec:foot_joints}
Our main results may suggest that the baseline outperforms other methods because of additional toe and heel joints. To identify the impact of these joints, we run the baseline with only 17 joints, removing toes and heels from the kinematic initialization.
Results are shown in Tables \ref{tab:results_baseline_supp} and \ref{tab:results_val_h36m_supp}.
In Table \ref{tab:results_baseline_supp}, we observe comparable plausibility metrics, but with noticeable improvements for GP and \metricTwoShort$_{100}$. These gains instead suggest that while the toes and heels provide more information about the orientation of the foot, it adds additional variance into the pose estimation. The observed physical plausibility improves when omitting the toe and heel joints.
In Table \ref{tab:results_val_h36m_supp}, we note similar per class performance to the baseline counterpart, with the most increases coming from the lower-performing classes that contain significant crouching or bending over movements.

\begin{table}[h]
\centering
    \begin{tabular}{l | l l l l}
    \toprule
        Method & FS (\%) & GP $\downarrow$ & \metricOneShort\ $\downarrow$ & \metricTwoShort$_{100}$ $\uparrow$ \\
    \midrule
        Baseline-17 & 1.6 & \textbf{0.11} & 28.9 & \textbf{74.7} \\
    \bottomrule
    \end{tabular}
    \caption{We show results on our validation subset on Human3.6M dataset. Baseline-17 removes the toe and heel keypoints. Results are comparable to Table 1 in the main text.}
    \label{tab:results_baseline_supp}
\end{table}

\begin{table*}[h]
    \centering
    \resizebox{0.95\textwidth}{!}{%
    \begin{tabular}{l|lllllllllll}
    \toprule
         Method & Dir. & Disc. &  Greet &  Photo &  Pose &  Purch. &  Wait &  WalkD. & WalkT. & Walk & Avg.\\
    \midrule
         Baseline-17 & 84.3&  75.7&  77.2&  \textbf{98.5} &  75.2 &  82.4 &  37.8 &  55.4 &  \textbf{83.3} & 77.0 & \textbf{74.7}\\
    \bottomrule
    \end{tabular}
    }
    \caption{Here we show the per-class performance for the \metricTwoShort$_{100}$ (Higher is better) metric on Baseline-17. Baseline-17 removes the toe and heel keypoints. Results are comparable to Table 2 in the main text.}
    \label{tab:results_val_h36m_supp}
\end{table*}

\subsection{Simulation details}
The CMA-ES algorithm is executed on $40$ cpus for 9-10 hours on $100$ frames in each video sequence. In a sliding fashion, two adjacent time windows are optimized for $200$ iterations. The result of the first window initializes the starting point of the next two subsequent time windows. We select the hyper-parameters in the cost function through manual fine-tuning to minimize the \metricOneShort\ trajectory distance.

\subsection{Qualitative Examples}
\begin{figure}[ht]
    \centering
    \includegraphics[width=0.95\textwidth]{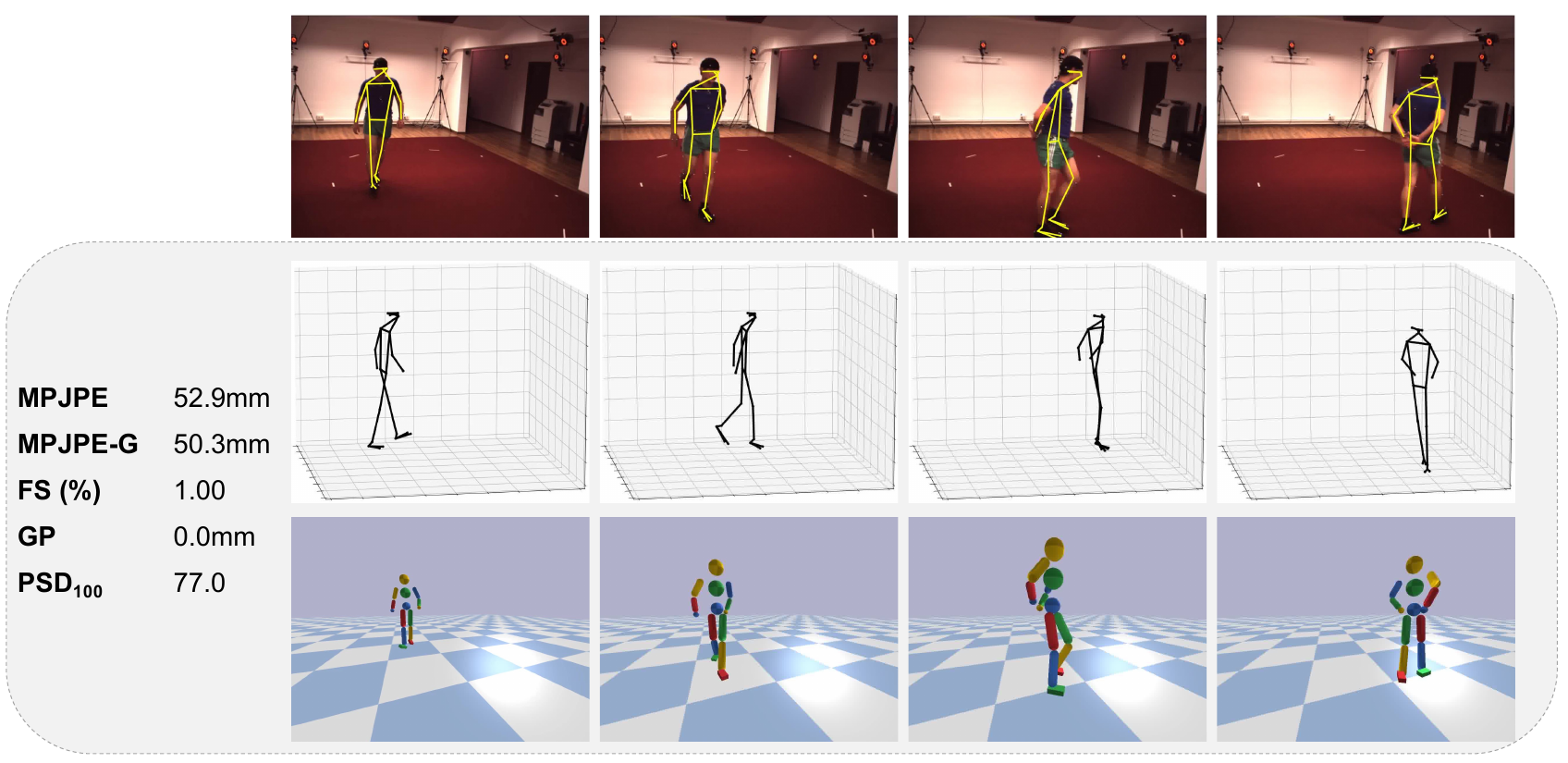}
    \caption{Results on the Baseline for the \textit{S11} - \textit{Walking 1} sequence.}
    \label{fig:supp_qual_003}
\end{figure}

\begin{figure}[ht]
    \centering
    \includegraphics[width=0.95\textwidth]{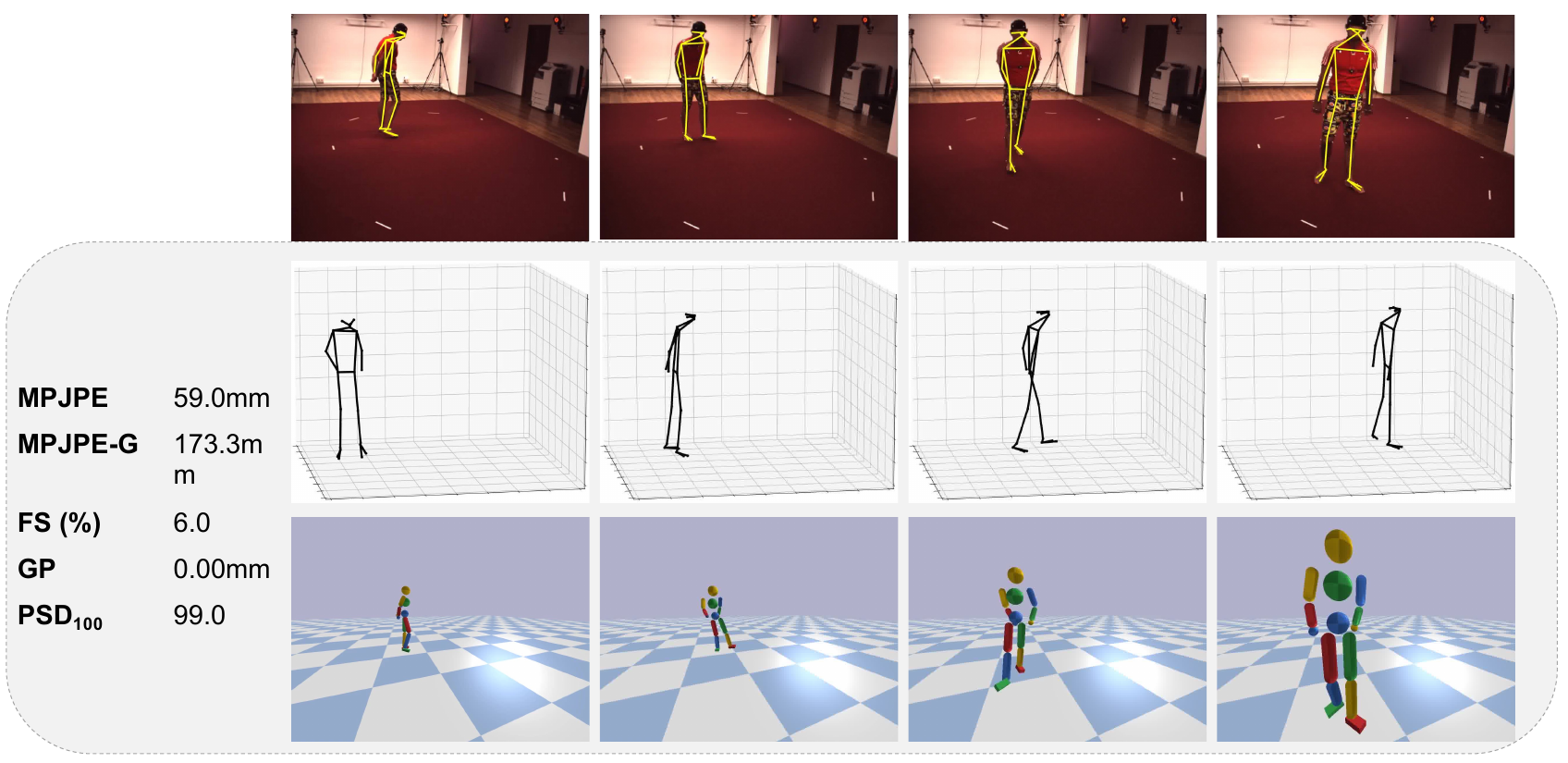}
    \caption{Results on the Baseline for the \textit{S9} - \textit{Waiting 1} sequence.}
    \label{fig:supp_qual_004}
\end{figure}

\begin{figure}[ht]
    \centering
    \includegraphics[width=0.95\textwidth]{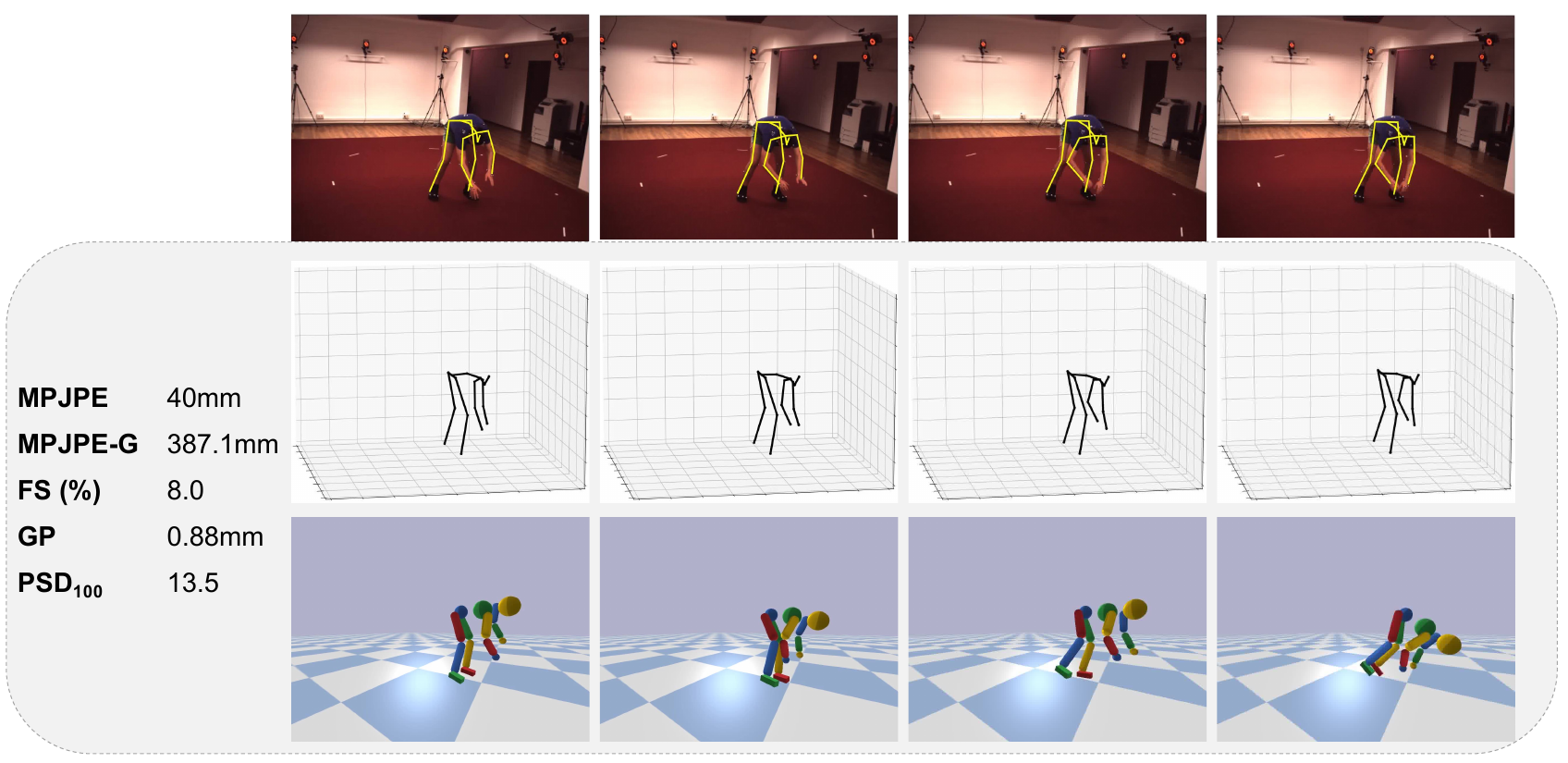}
    \caption{Results on PoseFormer for the \textit{S11} - \textit{Purchases 1} sequence.}
    \label{fig:supp_qual_001}
\end{figure}

\begin{figure}[ht]
    \centering
    \includegraphics[width=0.95\textwidth]{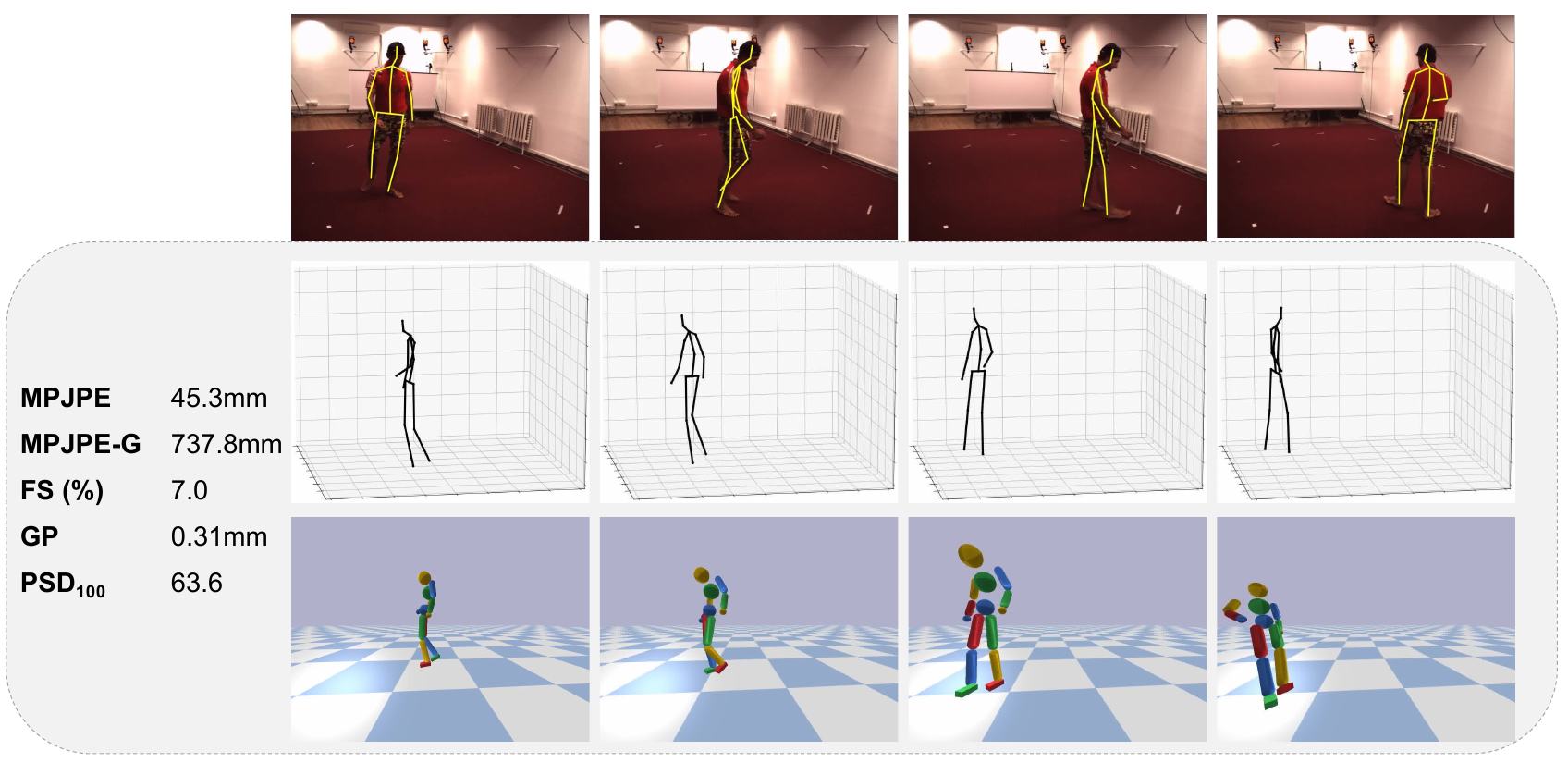}
    \caption{Results on PoseFormer for the \textit{S9} - \textit{WalkDog 1} sequence.}
    \label{fig:supp_qual_002}
\end{figure}

In Figures \ref{fig:supp_qual_003} and \ref{fig:supp_qual_004}, we show qualitative examples on two higher performing class, \textit{Walking} and \textit{Waiting}. 
We run both of these examples on our multi-view Baseline, where we note low MPJPE-G scores and no ground penetration. The consistent ground estimation, accurate multi-view estimation of the limbs, and the linear movement result in much more stable motion of the simulated body. 
In Figures \ref{fig:supp_qual_001} and \ref{fig:supp_qual_002}, we show qualitative examples on two of the lower performing classes, \textit{Purchases} and \textit{WalkDog}. We run both examples on the PoseFormer architecture. 
While MPJPE is low, the \textit{Purchases} sequence has higher ground penetration, suggesting an inconsistent estimation of the floor leading to instability. 
While we have observed that this is negligible with a stationary pose, the simulated body falls forward when bending over.
The \textit{WalkDog} sequence is more stable, but struggles when the simulated body does a turnaround, possibly due to sub-optimal optimization on ankle joint angle.

\clearpage
\bibliography{bmvc_final}